\documentclass[11pt,a4paper, dvipsnames]{article}
\usepackage{authblk}
\usepackage[hyperref]{emnlp-ijcnlp-2019}
\usepackage{times}
\usepackage{latexsym}
\usepackage{amsthm}
\usepackage{amsmath}
\usepackage{url}
\usepackage{natbib}
\usepackage{pgfplots}
\pgfplotsset{compat=1.14}
\usepackage{amsmath,amsfonts,bm}

\def\eqref#1{equation~\ref{#1}}
\def\1{\bm{1}}

\def\va{{\bm{a}}}

\def\ve{{\bm{e}}}

\def\vh{{\bm{h}}}

\def\vr{{\bm{r}}}

\def\vt{{\bm{t}}}

\DeclareMathAlphabet{\mathsfit}{\encodingdefault}{\sfdefault}{m}{sl}
\SetMathAlphabet{\mathsfit}{bold}{\encodingdefault}{\sfdefault}{bx}{n}

\def\sC{{\mathbb{C}}}

\newcommand{\R}{\mathbb{R}}

\usepackage{tabularx}
\usepackage{multirow}

\newcommand\blfootnote[1]{%
  \begingroup
  \renewcommand\thefootnote{}\footnote{#1}%
  \addtocounter{footnote}{-1}%
  \endgroup
}

\definecolor{figcolour1}{RGB}{27,158,119}
\definecolor{figcolour2}{RGB}{217,95,2}
\definecolor{figcolour3}{RGB}{117,112,179}
\definecolor{figcolour4}{RGB}{231,41,138}

\aclfinalcopy % Uncomment this line for the final submission
\title{Using Pairwise Occurrence Information to Improve Knowledge Graph Completion on Large-Scale Datasets}

\author[1,2*]{\textbf{Esma Balk{\i}r}}
\author[2]{\textbf{Masha Naslidnyk}}
\author[2]{\textbf{Dave Palfrey}}
\author[2]{\textbf{Arpit Mittal}}
\affil[1]{University of Edinburgh, Scotland, UK}
\affil[2]{Amazon Research, Cambridge, UK}
\affil[1]{\tt esma.balkir@ed.ac.uk}
\affil[2]{\tt \{naslidny, dpalfrey, mitarpit\}@amazon.co.uk}

\date{}
\begin{document}
\maketitle

\begin{abstract}
Bilinear models such as DistMult and \linebreak ComplEx are effective methods for knowledge graph (KG) completion. However, they require large batch sizes, which becomes a performance bottleneck when training on large scale datasets due to memory constraints. 
In this paper we use occurrences of entity-relation pairs in the dataset to construct a joint learning model and to increase the quality of sampled negatives during training. We show on three standard datasets that when these two techniques are combined, they give a significant improvement in performance, especially when the batch size and the number of generated negative examples are low relative to the size of the dataset. We then apply our techniques to a dataset containing 2 million entities and demonstrate that our model outperforms the baseline by 2.8\% absolute on hits@1.
\blfootnote{*Work done while the author was an intern}
\end{abstract}

\section{Introduction}

A Knowledge Graph (KG) is a collection of facts which are stored as triples, e.g. \textit{Berlin is-capital-of Germany}.
Even though knowledge graphs are essential for various NLP tasks, open domain knowledge graphs have missing facts. To tackle this issue, there has recently been considerable interest in KG completion methods, where the goal is to rank correct triples above incorrect ones. %For an overview, see \citet{Nickel2016AGraphs}.

Embedding methods such as 
%TransE \cite{Bordes2013TranslatingData} and
DistMult \cite{Yang2014EmbeddingBases} and 
ComplEx \cite{Trouillon2016ComplexPrediction} are simple and effective methods for this task, but are known to be sensitive to  hyperparameter and loss function choices \citep{Kadlec2017KnowledgeBack, Lacroix2018CanonicalCompletion}.
When paired with the right loss function, these methods need large minibatches and a large number of corrupted triples per each positive triple during training to reach peak performance. \footnote{The choice of loss function has a very strong effect on the optimal negative ratio, but with any loss function, larger batches tend to improve the results.} This causes memory issues for KGs in the wild, which are several magnitudes bigger than the common benchmarking datasets. 

%We break the batch size bottleneck by developing a flexible

To address the issue of scalability, we develop a framework that could be used with any bilinear KG embedding model.  We name our model \textbf{JoBi} (\textbf{Jo}int model with \textbf{Bi}ased negative sampling). Our framework uses occurrences of entity-relation pairs to overcome data sparsity, and to bias the model to score plausible triples higher. The framework trains a base model jointly with an auxiliary model that uses occurrences of pairs within a given triple in the data as labels. For example, the auxiliary model would receive the label 1 for the triple (\textit{Berlin is-capital-of France}) if the pairs (\textit{Berlin is-capital-of}) and (\textit{is-capital-of France}) are present in the training data, while the base model would receive the label 0. 

The intuition for using bigram occurrences is to capture some information about restrictions on the set of entities that could appear as the object or subject of a given relation; that information should implicitly correspond to some underlying type constraints. For example, even if \textit{(Berlin is-capital-of France)} is not a correct triple, \textit{Berlin} is the right type for the subject of \textit{is-capital-of}.

Our framework also utilizes entity-relation pair occurrences to improve the distribution of negative examples for contrastive training, by sampling a false triple e.g. (\textit{Berlin is-capital-of France}) with higher probability if the pairs (\textit{Berlin is-capital-of}) and (\textit{is-capital-of France}) both occur in the dataset. This tunes the noise distribution for the task so that it is more challenging, and hence the model needs a fraction of the negative examples compared to a uniform distribution.    

We show empirically that joint training is especially beneficial when the batch size is small, and biased negative sampling helps model learn higher quality embeddings with much fewer negative samples. We show that the two techniques are complementary and perform significantly better when combined. We then test JoBi on a large-scale dataset, and demonstrate that JoBi learns better embeddings in very large KGs.

% \begin{enumerate}
% \item Present a multi-task framework which \linebreak models both the correctness and the type consistency of each triple, and can be used to enrich any bilinear KG embedding model.
% \item Show that our framework significantly \linebreak improves on all baseline models when coupled with a negative sampling procedure that biases the sampling towards more informative negative samples.
% \item Analyze the effect of our framework on parameter settings that reflect training on large-scale datasets of different sizes.
% \item Apply the framework to a large-scale dataset, and demonstrate that the performance improves significantly.
% \end{enumerate}

\section{Background}

Formally, given a set of entities
$\mathcal{E} = \{e_0, \ldots, e_n\}$ and
a set of relations $\mathcal{R} = \{r_0, \ldots, r_m\}$, a Knowledge Graph (KG) is a set triples in the form $\mathcal{G} = \{(h, r, t)\} \subseteq \mathcal{E} \times \mathcal{R} \times \mathcal{E}$, where if a triple $(h, r, t) \in \mathcal{G}$, then relation $r$ holds between entities $h$ and $t$. Given such a KG, the aim of KG completion is score each triple in $\mathcal{E} \times \mathcal{R} \times \mathcal{E}$, so that correct triples are assigned higher scores than the false ones. KG embedding methods achieve this by learning dense vector representations for entities and relations through optimizing a chosen scoring function. 
A class of KG completion models such as RESCAL \citep{Nickel2012FactorizingData}, DistMult \citep{Yang2014EmbeddingBases}, ComplEx \citep{Trouillon2016ComplexPrediction}, SimplE \citep{Kazemi2018SimplEGraphs} and TUCKER \citep{balavzevic2019tucker} define their scoring function to be a bilinear interaction of the embeddings of entities and relations in the triple. For this work we consider DistMult, ComplEx and SimplE as our baseline models due to their simplicity.

\paragraph{DistMult.} \citep{Yang2014EmbeddingBases} is a knowledge graph completion model that defines the scoring function for a triple as a simple bilinear interaction, where the entity has the same representation regardless of whether it appears as the head or the tail entity. For entities $h, t$, relation $r$, and the embeddings $\vh, \vt, \vr \in \R^d$, the scoring function is defined as:
\begin{equation} \label{scorefun}
s(h, r, t) = \vh^T \, \text{diag}(\vr) \, \vt
\end{equation}

 where  $\text{diag}(\vr)$ is a diagonal matrix with $\vr$ on the diagonal.
 
 \paragraph{ComplEx.} \citep{Trouillon2016ComplexPrediction} is a bilinear model similar to DistMult. Because of its symmetric structure, DistMult cannot model anti-symmetric relations. ComplEx overcomes this shortcoming by learning embeddings in a complex vector space, and defining the embedding of an entity in tail position as the complex conjugate of the embedding in the head position.
 
 Let $\vh, \vr, \vt \in \sC^d$ be the embeddings for $h, r, t$. The score for ComplEx is defined as follows:
\begin{equation}
s(h, r, t) = \text{Re}\left(\vh^T \text{diag}(\vr)  \overline{\vt}\right)
\end{equation}

Where $\overline{\va}$ denotes the complex conjugate of $\va$, and $\text{Re}(\va)$ denotes the real part of the complex vector $\va$.

\paragraph{SimplE.} \citep{Kazemi2018SimplEGraphs} is also a bilinear model similar to DistMult. Each entity has two associated embeddings $\ve_1, \ve_2 \in \R^d$, where one is the representation of $e$ as the head, and the other as the tail entity of the triple. Each relation also has two associated embeddings: $\vr$ and $\vr^{-1}$, where $\vr^{-1}$ is the representation for the reverse of $r$. The score function is defined as: 
\begin{align}
\begin{split}
 s(h, r, t) = 1/2  \left(\vh_{1} \right)^T \text{ diag}(\vr) \vt_1 \\ + 1/2 \left(\vt_2\right)^T \text{ diag}(\vr^{-1}) \vh_2
% s(e_1, r, e_2) = \frac{1}{2}  \left(\ve_1^h \right)^T \text{ diag}(\vr) \ve_2^t  \\ + \frac{1}{2} \left(\ve_2^h\right)^T \text{ diag}(\vr^{-1}) \ve_1^t  
\end{split}
\end{align}

 \section{Joint framework} JoBi contains two copies of a bilinear model, where one is trained on labels of triples, and the other on occurrences of entity-relation pairs within the triples. For the pair module, we label a triple $(h,r,t)$ correct if there are triples $(h, r, t')$ and $(h', r, t)$ in the training set for some $t'$ and $h'$. 
 
 The scoring functions for the two models are $s_{\text{bi}}$ and $s_{\text{tri}}$, for the pair and the triple modules respectively. We tie the weights of the entity embeddings, but let the embeddings for the relations be optimized separately. The equations using ComplEx as the base model are as follows:
 \begin{align}
    s_{\text{tri}}(h, r, t) &= \text{Re}\left(\vh^T \text{diag}(\vr_{\text{tri}})  \overline{\vt}\right) \\
    s_{\text{bi}}(h, r, t) &= \text{Re}\left(\vh^T \text{diag}(\vr_{\text{bi}})  \overline{\vt}\right) 
\end{align}

We define the framework for DistMult and SimplE analogously. During training, we optimize the two jointly, but use only $s_\text{tri}$ during test time. Hence, the addition of the auxiliary module has no effect on the number of final parameters of the trained model. Note that even during training, this doesn't increase model complexity in any significant way since the number of relations in KGs are often a fraction of the number of entities. 
 
 For each triple in the minibatch, we generate $n_{\text{neg}}$ negative examples per positive triple by randomly corrupting the head or the tail entity. For $s_{\text{tri}}$, we use the negative log-likelihood of softmax as the loss function, and for $s_{\text{bi}}$ we use binary cross entropy loss. We combine the two losses via a simple weighted addition with a tunable hyperparameter $\alpha$:
 \begin{equation}
     \mathcal{L}_\text{total} = \mathcal{L}_\text{tri} + \alpha\mathcal{L}_\text{bi}
 \end{equation}
 
 \paragraph{Biased negative sampling.} We also examine the effect of using the pair cooccurrence information for making the contrastive training more challenging for the model. For this, we keep the model as is, but with probability $p$, instead of corrupting the head or the tail of the triple with an entity chosen uniformly at random, we corrupt it with an entity that is picked with uniform probability from the set of entities that occur as the head or tail entity of the relation in the given triple. To illustrate, when sampling a negative tail entity for the tuple \textit{Berlin is-capital-of}, this method causes the model to pick \textit{France} with higher probability than \textit{George Orwell} if \textit{France} but not \textit{George Orwell} occurs as the head entity for the relation \textit{is-capital-of} in the training data. 

 \section{Experiments} \label{sec:experiments}

  \begin{table}[t!] % [H]
    \centering
    \small
    \begin{tabular}{r | c c c}
          \textbf{Dataset} & \textbf{\#entities} & \textbf{\#relations} & \textbf{\#train} \\
         \hline 
        FB15K & 14,951 & 1,345 & 483,142 \\
        \hline
        FB15K-237 & 14,541 & 237 & 272,115 \\
        \hline
        YAGO3-10 & 123,182 & 37 & 1,079,040 \\
        \hline
        FB1.9M & 1,892,241 & 3,247 & 19,323,513 \\
    \end{tabular}
    \caption{Statistics of datasets used in experiments.}
    \label{tab:datasets}
\end{table}

We perform our experiments on standard datasets FB15K \citep{Bordes2013TranslatingData}, FB15K-237 \citep{Toutanova2015RepresentingBases}, YAGO3-10 \citep{Dettmers2018ConvolutionalEmbeddings}, and on a new large-scale dataset FB1.9M which we construced from FB3M \citep{Xu2018InvestigationsExtraction}. \footnote{We do not perform experiments on WordNet derived datasets WN18 or WN18RR because bigram modelling would not provide any information -- all entities are synsets and almost all can occur as an object or subject to all the possible relations.} We focus on YAGO3-10 since it is 10 times larger than the other two and better reflects how the performance of the models scale. We present the comparison of the sizes of these datasets in Table \ref{tab:datasets}, and further details could be found in Appendix \ref{sec:Datasets}. 

For evaluation, we rank each triple $(h, r, t)$ in the test set
against $(h', r, t)$ for all entities $h'$, and similarly against $(h, r, t')$ for all entities $t'$. We filter out the candidates that have occurred in training, validation or test set as described in \citet{Bordes2013TranslatingData}, and we report average hits@1, 3, 10 and mean reciprocal rank (MRR). 
 
% We re-implement all our baselines and optimize them to obtain very competitive results. A table comparing our baselines to other implementations is provided in Appendix \ref{sec:BaselinesComparison}. The code will be made publicly available.

We re-implement all our baselines and obtain very competitive results. In our preliminary experiments on baselines, we found that the choice of loss function had a large effect on performance, with negative log-likelihood (NLL) of softmax consistently outperforming both max-margin and logistic losses. Larger batch sizes lead to better performance. With NLL of sampled softmax, we found that increasing the number of generated negatives steadily increases performance\footnote{This effect is not observed when using logistic-loss or max-margin loss}, and state-of-the-art results could be reached by using the full softmax as used in \citet{Joulin2017FastEmbeddings} and \citet{Lacroix2018CanonicalCompletion}. This technique is possible for standard benchmarks but not for large KGs, and we report results in Appendix \ref{sec:full-softmax} 
for all datasets small enough to allow for full contrastive training. However, our main experiments use NLL of sampled softmax since our focus is on scalability. Note that results with full softmax (Appendix \ref{sec:full-softmax}) demonstrate that our implementation of baselines is very competitive.Our implementation of ComplEx performs significantly better than ConvE \citep{Dettmers2018ConvolutionalEmbeddings} on two out of the three datasets, and come close to results of \citet{Lacroix2018CanonicalCompletion} who use extremely large embeddings as well as full softmax, thus cannot be scaled. Our code is publicly available. \footnote{\tiny{{\tt  https://github.com/awslabs/joint\_biased\_embeddings}}}

For most of our experiments, we choose to use ComplEx as the base for our model (\textbf{JoBi ComplEx}), since this configuration consistently outperformed others in preliminary experiments. To test the effect of our techniques on different bilinear models, we report results with DistMult (\textbf{JoBi DistMult}) and SimplE (\textbf{JoBi SimplE}) on FB15K-237. 
% Since \citet{Jain2017JointInference} report results using a non-standard evaluation method, partial comparison is offered in Appendix \ref{sec:typecomplex}. 

\begin{table}[t!] % [H]
    \centering
    \small
    \begin{tabular}{r | c c c c  }
     \hline \hline
      \multicolumn{1}{l|}{\textbf{FB15K-237}}  & \textit{h@1} & \textit{h@3} & \textit{h@10} & \textit{MRR} \\
         \hline
         SimplE & 0.160 & 0.268 & 0.430 & 0.248 \\
         DistMult & 0.158 & 0.271 & 0.432 & 0.247  \\
         ComplEx & 0.159 & 0.275 & 0.441 &0.25 \\ 
         JoBi SimplE  & 0.188 & 0.301 & 0.461 & 0.277 \\
         JoBi DistMult  & \textbf{0.205} & 0.316 & 0.466 & 0.29 \\
         JoBi ComplEx  & 0.199 & \textbf{0.319} & \textbf{0.479} & \textbf{0.29}  \\
         \hline \hline
       \multicolumn{1}{l|}{\textbf{FB15K}}  & \textit{h@1} & \textit{h@3} & \textit{h@10} & \textit{MRR} \\
         \hline 
         DistMult & 0.587 & 0.785 & 0.867 & 0.697 \\
         ComplEx & 0.617 & 0.803  & 0.874 & 0.72 \\ 
         JoBi ComplEx  & \textbf{0.681} & \textbf{0.824} & \textbf{0.883} & \textbf{0.761}  \\
         \hline \hline
       \multicolumn{1}{l|}{\textbf{YAGO3-10}}  & \textit{h@1} & \textit{h@3} & \textit{h@10} & \textit{MRR} \\
         \hline
         DistMult & 0.252 & 0.407 & 0.568 & 0.357 \\
         ComplEx & 0.277 & 0.44 & 0.589 & 0.383 \\ 
         JoBi ComplEx  & \textbf{0.333} & \textbf{0.477} & \textbf{0.617} & \textbf{0.428}  \\
         
    \end{tabular}
    \caption{Performance on different datasets against baselines, where h@$k$ denotes hits at $k$. Results are reported on test sets with the best parameters found in grid search for each model.}
    \label{tab:smalldatasets}
\end{table}

\paragraph{Discussion.} It could be seen in Table \ref{tab:smalldatasets} that JoBi ComplEx outperforms both ComplEx and DistMult on all three standard datasets, on all the metrics we consider. For Hits@1, JoBi Complex outperforms baseline ComplEx by 4\% on FB15K-237, 6.4\% on FB15K and 5.6\% on YAGO3-10. 

Moreover, results in Table \ref{tab:smalldatasets} demonstrate that JoBi improves performance on DistMult and SimplE. It should be noted that on FB15K-237, all JoBi models outperform all the baseline models, regardless of the base model used.

Lastly, results on FB1.9M (Table \ref{tab:fb2p6m}) demonstrate that JoBi improves performance on this very large dataset, where it is not possible to perform softmax over the entire set of entities, or have very large embedding sizes due to memory constraints. 

\begin{table}[t!] % [H]
    \centering
    \small
    \begin{tabular}{r | c c c c  }
       \textbf{ComplEx} & \textit{h@1} & \textit{h@3} & \textit{h@10} & \textit{MRR} \\
         \hline 
         Baseline & 0.424 & 0.598 & 0.721 & 0.530 \\
         JoBi  & \textbf{0.452} & \textbf{0.615} & \textbf{0.726} & \textbf{0.550}

    \end{tabular}
    \caption{Performance on the large-scale FB1.9M dataset, measured against the best performing baseline.}
    \label{tab:fb2p6m}
\end{table}

Although one epoch for JoBi takes slightly longer than the baseline, JoBi converges in fewer epochs, resulting in shorter running time overall. We report running times on FB1.9M in Table \ref{tab:runtime}.

\begin{table}[t!]
    \centering
    \small
    \begin{tabular}{c | c | c}
    & \textbf{\# epochs} & \textbf{training time} \\
    \hline
    ComplEx & 70 & 5 days 5 hours 8 minutes \\
    JoBi ComplEx & 30 & 4 days 19 minutes 
    \end{tabular}
    \caption{Runtimes of ComplEx and JoBi Complex on FB1.9M.}
    \label{tab:runtime}
\end{table}

\paragraph{Comparison with TypeComplex} \label{sec:typecomplex}
For results of TypeComplex, \citet{Jain2018Type-SensitiveSupervision} use a wider set of negative ratios in their grid search than we do. To isolate the effects of the different models from hyperparameter choices, we set the negative ratio for our model to be 400 to match the setting on their best performing models. We keep the other hyperparameters the same as the best performing models for the previous experiments. 

\citet{Jain2018Type-SensitiveSupervision} use a modified version of the ranking evaluation procedure to report their results, where they only rank the tail entity against all other entities. To be able to compare our model to theirs, we also report the performance of our framework on this modified metric. The results for these experiments can be found in Table \ref{tab:typecomplex}.

\begin{table}
    \centering
    \small
    \begin{tabular}{r | c c c c  }
      \multicolumn{1}{l|}{\textbf{FB15K-237}}  & \textit{h@1} & \textit{h@3} & \textit{h@10} & \textit{MRR} \\
         \hline 
         ComplEx & 0.209 & 0.347 & 0.535 & 0.314 \\
         TypeComplex & \textbf{0.296} & - & 0.575 & \textbf{0.389} \\ 
         JoBi ComplEx  & 0.276 & 0.416  & \textbf{0.587}  & 0.377 \\
         \hline \hline  
       \multicolumn{1}{l|}{\textbf{FB15K}}  & \textit{h@1} & \textit{h@3} & \textit{h@10} & \textit{MRR} \\
         \hline 
          ComplEx & 0.630 & 0.818 & 0.895 & 0.734 \\
         TypeComplex & 0.663 & - & 0.885 & 0.754 \\ 
         JoBi ComplEx  & \textbf{0.702} & 0.847 & \textbf{0.906} & \textbf{0.782} \\
         \hline\hline
       \multicolumn{1}{l|}{\textbf{YAGO3-10}}  & \textit{h@1} & \textit{h@3} & \textit{h@10} & \textit{MRR} \\
         \hline
        ComplEx & 0.412 & 0.587 & 0.701 & 0.516 \\
         TypeComplex & \textbf{0.516} & - & 0.702 & 0.587 \\ 
         JoBi ComplEx  & 0.507 & 0.647 & \textbf{0.742}  & \textbf{0.591}
    \end{tabular}
    \caption{Comparison with TypeComplex where the scores are calculated ranking only the tail entities. Results for TypeComplex are taken from \citet{Jain2018Type-SensitiveSupervision}. h@$k$ denotes hits at $k$.}
    \label{tab:typecomplex}. 
\end{table}

Our model generally outperforms TypeComplex by a large margin on hits@10. It also outperforms TypeComplex on MRR by a moderate margin except on FB15K-237, the smallest dataset. On the other hand, TypeComplex outperforms our model on hits@1 in two out of the three datasets. In fact for FB15K, TypeComplex does worse on hits@10 compared to the baseline model. This suggests that TypeComplex may be compromising on hits@k where k is larger to improve the hits@1 metric, which might be undesirable depending on the application. 

\paragraph{Qualitative analysis.} We analyzed correct predictions made by JoBi ComplEx but not regular ComplEx. Among relations in YAGO3-10, major gains can be observed for \textit{hasGender} (Appendix \ref{sec:gains}). The improvement comes solely from tail-entity predictions, with hits@1 increasing from 0.22 to 0.86. Furthermore, we found that the errors made by ComplEx are exactly of the kind that can be mitigated by enforcing plausibility: ComplEx predicts an object that is not a gender (e.g. a sports team or a person) 65\% of the time; JoBi makes such an obvious mistake only 2\% of the time.

\paragraph{Ablation studies.} We compare joint training without biased sampling (\textbf{Joint}) and biased sampling without joint training (\textbf{BiasedNeg}) to the full model JoBi on YAGO3-10. The results can be found in Table \ref{tab:results}.  We also conduct experiments to isolate the effect of our techniques on varying batch sizes and negative ratios. The results for this experiment are presented in Figures \ref{fig:results1} and \ref{fig:results2}.  Training details can be found in Appendix \ref{sec:implementation}. 

In Table \ref{tab:results} it can be seen that Joint on its own gives a slight performance boost over the baseline, and BiasedNeg performs slightly under the baseline on all measures. However, combining our two techniques in JoBi gives 5.6\% points improvement on hits@1. This suggests that biased negative sampling increases the efficacy of joint training greatly, but is not very effective on its own.

\begin{table}
    \centering
    \small
    \begin{tabular}{r | c c c c}
         & \textit{h@1} & \textit{h@3} & \textit{h@10} & \textit{MRR} \\
         \hline %\hline
         Baseline & 0.277 & 0.44 & 0.589 & 0.383 \\
         BiasedNeg & 0.276 & 0.427 & 0.568  & 0.375 \\ 
          Joint & 0.287 & 0.447 & 0.601 & 0.392 \\
         JoBi  & \textbf{0.333} & \textbf{0.477} & \textbf{0.617} & \textbf{0.428}  \\
         \hline
         
    \end{tabular}
    \caption{Results of ablation study on ComplEx model.}
    %Results are reported on the test set with the best parameters found in grid search for each model. }
    \label{tab:results}
\end{table}

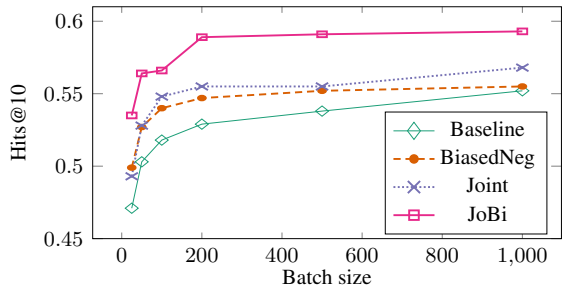
\begin{figure}[t!]%[H]
\begin{tikzpicture}[font=\small, scale=0.9]

\begin{axis}[
xlabel={Batch size},
ylabel={Hits@10},
ymin=0.45, ymax=0.61,
legend pos=south east,
yscale=0.6
]
\addplot+[color=figcolour1, mark=diamond, mark options={scale=1.8}] coordinates 
%checked
{
(25, 0.471)
(50, 0.503)
(100, 0.518)
(200, 0.529)
(500, 0.538)
(1000, 0.552)
};
\addlegendentry{Baseline}

\addplot+[color=figcolour2, style={thick, densely dashed}, mark=*, mark options={solid, scale=0.8}] coordinates 
%checked
{
(25, 0.499)
(50, 0.527)
(100, 0.540)
(200, 0.547)
(500, 0.552)
(1000, 0.555)
};

\addlegendentry{BiasedNeg}
\addplot+[color=figcolour3, mark=x, mark options={solid, scale=1.8}, style={thick, densely dotted}] coordinates 
%checked
{
(25, 0.493)
(50, 0.528)
(100, 0.548)
(200, 0.555)
(500, 0.555)
(1000, 0.568)
};

\addlegendentry{Joint}
\addplot+[color=figcolour4, mark=square, mark options={scale=1}, style={thick}] coordinates 
%checked
{
(25, 0.535)
(50, 0.564)
(100, 0.566)
(200, 0.589)
(500, 0.591)
(1000, 0.593)
};

\addlegendentry{JoBi}
\end{axis}
\end{tikzpicture} \caption{Performances on YAGO3-10 with different batch sizes} 
\label{fig:results1}
\end{figure}

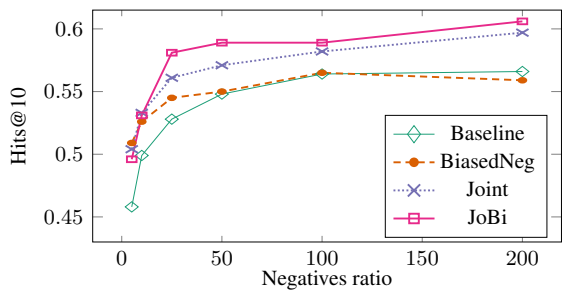
\begin{figure}[t!]
\begin{tikzpicture}[font=\small, scale=0.9]

\begin{axis}[
xlabel={Negatives ratio},
ylabel={Hits@10},
ymin=0.43, ymax=0.615,
legend pos=south east,
yscale=0.6
]
\addplot+[color=figcolour1, mark=diamond, mark options={scale=1.8}]  coordinates 
%checked
{
(5, 0.458)
(10, 0.499)
(25, 0.528)
(50, 0.548)
(100, 0.564)
(200, 0.566)
};
\addlegendentry{Baseline}

\addplot+[color=figcolour2, style={thick, densely dashed}, mark=*, mark options={solid, scale=0.8}]  coordinates 
%checked
{
(5, 0.509)
(10, 0.526)
(25, 0.545)
(50, 0.550)
(100, 0.565)
(200, 0.559)
};
\addlegendentry{BiasedNeg}

\addplot+[color=figcolour3, style={thick, densely dotted}, mark=x, mark options={solid, scale=1.8}] coordinates 
%checked
{
(5, 0.504)
(10, 0.533)
(25, 0.561)
(50, 0.571)
(100, 0.582)
(200, 0.597)
};
\addlegendentry{Joint}

\addplot+[color=figcolour4, mark=square, mark options={scale=1}, style={thick}] coordinates 
%checked
{
(5, 0.496)
(10, 0.531)
(25, 0.581)
(50, 0.589 )
(100, 0.589 )
(200, 0.606)
};
\addlegendentry{JoBi}

\end{axis}
\end{tikzpicture}
\caption{Performances on YAGO3-10 with different negative ratios} 
\label{fig:results2}
\end{figure}

Figure \ref{fig:results1} and \ref{fig:results2} shows that JoBi not only consistently performs the best over the entire range of parameters, but also delivers a performance improvement that is especially large when the batch size or the negative ratio is small. This setting was designed to reflect the training conditions on very large datasets. It can be seen that BiasedNeg is more robust to low values of negative ratios, and both BiasedNeg and Joint alone show less deterioration in performance as the batch size decreases. When these two methods are combined in JoBi, the training becomes more robust to different choices on both these parameters.

The reason behind BiasedNeg performing worse on its own but better with Joint could be the choice of binary cross entropy loss for the pair module. 
%the two techniques complement each other especially well might be related to using binary cross entropy (BCE) loss for the pair module in Joint. 
We speculate that as the negative ratio increases, the ratio of negative to positive examples for this module becomes more skewed. Biasing the negative triples in the training alleviates this problem by making the classes more balanced, and allows the joint training to be more effective.

 \subsection{Related work} Utilizing pair occurrences for embedding models have been considered before, both as explicit model choices and as negative sampling strategies. \citet{chang2014typed} and \citet{Krompa2015Type-constrainedGraphs} use pair occurrences to constrain the set of triples to be used in the optimization procedure.
 For methods that rely on SGD with contrastive training, this translates to a special case of our biased sampling method where $p = 1$.  
 \citet{Garcia-Duran2016CombiningBases} present \textit{TATEC}, a model that combines bigram and trigram interactions. 
% Note: I didn't change the bigram/trigram wording here on purpose because that is how they describe it in the paper.
 The trigram model uses a full matrix representation for relations, and hence has many more parameters compared to our model. \citet{Jain2018Type-SensitiveSupervision} present \textit{JointDM} and \textit{JointComplex}, which could be viewed as a simplification of \textit{TATEC}.
%  where the relation representation for the trigram term is either DistMult or ComplEx, and the bigram term 
%  that captures the similarity between the head and the tail entity is discarded.
Unlike our model, both of these methods use the bigram terms both in training and evaluation, do not share any of the embeddings between two models, and do not provide supervision based on pair occurrences in the data. Other methods that have been considered for improving the negative sampling procedure includes adversarial \citep{cai2018kbgan} and self-adversarial \citep{sun2018rotate} training. None of these methods focus on improving the models to scale to large KGs. 

\section{Conclusion}

We have presented a joint framework for KG completion that utilizes entity-relation pair occurrences as an auxiliary task, and combined it with a technique to generate informative negative examples with higher probability. We have shown that joint training makes the model more robust to smaller batch sizes, and biased negative sampling to different values of the number of generated negative samples. Furthermore, these techniques perform well above baselines when combined, and are effective on a very large KG dataset. Applying JoBi to non-bilinear models is also possible, but left for future work.

\bibliography{references}
\bibliographystyle{acl_natbib}
\newpage

\appendix

\section{Datasets} \label{sec:Datasets}
 FB15K \citep{Bordes2013TranslatingData} is a dataset derived from Freebase. FB15K-237 \citep{Toutanova2015RepresentingBases} is a subset of FB15K which only contains the most frequent 237 relations, and where the inverse relations are removed to prevent test leakage.  YAGO3-10 \citep{Dettmers2018ConvolutionalEmbeddings} is a dataset derived from YAGO-3 \citep{Suchanek2007Yago}, where each entity occurs with at least 10 relations. FB1.9M is a large-scale dataset we have constructed from FB3M \citep{Xu2018InvestigationsExtraction}, a large dataset derived from Freebase by iteratively removing entities that occur in less than 5
triples until no such entities remain. The statistics for each of these datasets could be found in Table \ref{tab:datasets}.

\section{Implementation details}\label{sec:implementation}
We optimize all models with stochastic gradient descent using Adam \citep{Kingma2014Adam:Optimization}, and perform early stopping with hits@10 on the validation set, where evaluation is performed every five epochs. For all our experiments, we fix initial learning rate to 0.001.

For experiments on FB15K, FB15K-237 and YAGO3-10, we fix embedding size to be 200. While performance increases with embeddings up to 2000 dimensions \cite{Kadlec2017KnowledgeBack, Lacroix2018CanonicalCompletion}, we cap ours at 200 to emulate constraints faced when dealing with very large KGs.  We perform a grid search over batch sizes: $\{500, 1000\}$, negative ratios $ n_{neg}: \{50, 100\}$, pair-loss weight $\alpha$ : $\{0.5, 1\}$  where applicable, and fix biased sampling probability $p$ to $0.3$. We choose the hyperparameters that give the highest hits@10 on the validation set, and use these hyperparameters to report the final results on the test set. 
%The results are presented in Table \ref{tab:smalldatasets}.

%   \begin{table}[t!] % [H]
%     \centering
%     \small
%     \begin{tabular}{r | c c c}
%           \textbf{Dataset} & \textbf{\#entities} & \textbf{\#relations} & \textbf{\#train} \\
%          \hline 
%         FB15K & 14,951 & 1,345 & 483,142 \\
%         \hline
%         FB15K-237 & 14,541 & 237 & 272,115 \\
%         \hline
%         YAGO3-10 & 123,182 & 37 & 1,079,040 \\
%         \hline
%         FB1.9M & 1,892,241 & 3,247 & 19,323,513 \\
%     \end{tabular}
%     \caption{Statistics of datasets used in experiments.}
%     \label{tab:datasets}
% \end{table}

For FB1.9M, we use the best hyperparameters from YAGO3-10. Due to memory constraints, we set the embedding size to 100. %Results for this dataset are presented in Table \ref{tab:fb2p6m}.

For the ablation study on YAGO3-10, we perform a grid search over batch sizes: $\{200, 500, 1000\}$, negative ratios $ n_{neg}: \{50, 100\}$, biased sampling probability $p: \{0.1, 0.2, 0.3\}$ and pair-loss weight $\alpha: \{0.25, 0.5\}$ where applicable. 

For demonstrating how the effects of JoBi compares to baselines with varying batch sizes, we keep everything but the batch size constant ($n_{neg}=25$, $\alpha= 0.5$, $p= 0.3$) and plot the change in hits@10 as the batch size varies in $\{25, 50, 100, 200, 500, 1000\}$. For demonstrating the effects of varying negative ratios, we keep everything but $n_{\text{neg}}$ constant (batch size = 200, $\alpha= 0.5$, $p= 0.3$) and plot hits@10 as $n_{\text{neg}}$ in $\{5, 10, 25, 50, 100, 200\}$.

\section{Qualitative comparison between ComplEx and JoBi ComplEx} \label{sec:gains}
\begin{figure}[h]
\includegraphics[width=8cm]{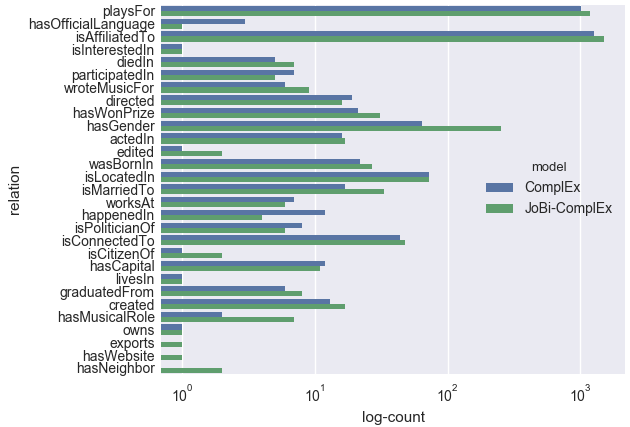}
\caption{Number of correctly predicted entities (hits@1) for ComplEx and JoBi ComplEx, broken down by relation.} 
\label{fig:gains}
\end{figure}

\section{Experiments with full-softmax}\label{sec:full-softmax}
Although the main focus of our framework is scalable training methods that use sampled negatives, we also test our joint method with softmax over the entire set of entities, and report the results in Table \ref{tab:fullSM}.
We can see that joint training improves the performance also when used with full-softmax on all the datasets,
beating a sophisticated, high performing method such as ConvE \cite{Dettmers2018ConvolutionalEmbeddings}, and coming close to the performance of ComplEx-N3 \citep{Lacroix2018CanonicalCompletion} which uses embeddings 10 times larger in size than ours. We note that neither full-softmax, nor embedding sizes used by \citet{Lacroix2018CanonicalCompletion} are scalable to large datasets. 

\begin{table} % [H]
    \centering
    \small
    \begin{tabular}{r | c c c c  }
      \multicolumn{1}{l|}{\textbf{FB15K-237}}  & Hits@1 & Hits@3 & Hits@10 & MRR \\
         \hline 
         SimplE & 0.227 & 0.340 & 0.492 & 0.314 \\
         DistMult &  0.225 & 0.343 & 0.490 & 0.313  \\
         ComplEx &  0.229 & 0.348 & 0.502 & 0.319 \\
         ConvE*   & 0.237 & 0.356 & 0.501 & 0.325\\
         ComplEx-N3\textdagger  & - & - & 0.56 & 0.37 \\
         JoBi ComplEx & 0.238 & 0.357 & 0.509 & 0.327 \\
         \hline \hline
       \multicolumn{1}{l|}{\textbf{FB15K}}  & Hits@1 & Hits@3 & Hits@10 & MRR \\
         \hline 
         DistMult & 0.779 & 0.844 & 0.890 & 0.819   \\
         ComplEx & 0.805 & 0.859 & 0.899 & 0.839   \\
         ConvE*   & 0.558 & 0.723 &  0.831 & 0.657 \\
         ComplEx-N3\textdagger  & - & - & 0.91 &  0.86  \\
         JoBi ComplEx & 0.804 & 0.861 & 0.901 & 0.840 \\
         \hline \hline
       \multicolumn{1}{l|}{\textbf{YAGO3-10}}  & Hits@1 & Hits@3 & Hits@10 & MRR \\
         \hline
         DistMult & 0.451 & 0.583 & 0.683 & 0.534  \\
         ComplEx & 0.468 & 0.599 & 0.702 & 0.550 \\
         ConvE*  & 0.35 & 0.49 & 0.62 &  0.44 \\
         ComplEx-N3\textdagger  & - & - &  0.71 & 0.58 \\
         JoBi ComplEx & 0.473 & 0.599 & 0.695 & 0.552  \\
    \end{tabular}
    \caption{Performance on different datasets against baselines and state-of-the-art methods using full-softmax. *\citet{Dettmers2018ConvolutionalEmbeddings} \textdagger\citet{Lacroix2018CanonicalCompletion}}
    \label{tab:fullSM}
\end{table}

\bibliography{references}
\bibliographystyle{acl_natbib}

\end{document}